\documentclass[conference]{IEEEtran}
\usepackage{cite}

\ifCLASSINFOpdf
  \usepackage[pdftex]{graphicx}
  \graphicspath{{fig/}}
  \DeclareGraphicsExtensions{.pdf}
\else
\fi
\usepackage{url}

\begin{document}
\bstctlcite{nourl}

%
\title{Models of human behavior for human-robot interaction and automated driving: How accurate do the models of human behavior need to be?}

\author{\IEEEauthorblockN{Gustav Markkula}
\IEEEauthorblockA{Institute for Transport Studies\\
University of Leeds\\
Leeds, United Kingdom\\
Email: g.markkula@leeds.ac.uk}
\and
\IEEEauthorblockN{Mehmet Dogar}
\IEEEauthorblockA{School of Computing\\
University of Leeds\\
Leeds, United Kingdom\\
Email: m.r.dogar@leeds.ac.uk}
}


%

\IEEEspecialpapernotice{This paper has been accepted for publication in IEEE Robotics \& Automation Magazine: \protect{\url{https://doi.org/10.1109/MRA.2022.3182892}}}

\maketitle

\begin{abstract}
There are many examples of cases where access to improved models of human behavior and cognition has allowed creation of robots which can better interact with humans, and not least in road vehicle automation this is a rapidly growing area of research. Human-robot interaction (HRI) therefore provides an important applied setting for human behavior modeling – but given the vast complexity of human behavior, how complete and accurate do these models need to be? Here, we outline some possible ways of thinking about this problem, starting from the suggestion that modelers need to keep the right end goal in sight: A successful human-robot interaction, in terms of safety, performance, and human satisfaction. Efforts toward model completeness and accuracy should be focused on those aspects of human behavior to which interaction success is most sensitive. We emphasise that identifying which those aspects are is a difficult scientific objective in its own right, distinct for each given HRI context. We propose and exemplify an approach to formulating a priori hypotheses on this matter, in cases where robots are to be involved in interactions which currently take place between humans, such as in automated driving. Our perspective also highlights some possible risks of overreliance on machine-learned models of human behavior in HRI, and how to mitigate against those risks.
\end{abstract}


%
\IEEEpeerreviewmaketitle

\section{Introduction}
Human behavior does not cease to fascinate: While clearly rule-bound and predictable in many ways, it is also consistently variable, adaptive, individual, or even (seemingly?) random. Unpeeling the layers of this complexity to better understand human behavior, and the cognitive processes underlying it, remains one of the most challenging and intriguing goals of science. For this reason, researchers across many disciplines (sociology, psychology, biology, etc.) develop \emph{models} of what human behavior looks like and why, at a range of different levels of abstraction, from conceptual to computational, and from mechanistic to machine-learned \cite{farrell_computational_2018,moustafa_computational_2017}.

Besides this push from fundamental scientific curiosity, there is also a pull, in the form of various applied uses for such models, one of them clearly in human-robot interaction (HRI) \cite{thomaz_computational_2016}. The HRI literature is full of examples where models of human behavior or cognition have been put to good use in order to improve the design of robotic agents interacting with humans \cite{thomaz_computational_2016,strabala_towards_2013,babel_step_2022,huang_adaptive_2015,hajiseyedjavadi_effect_2021,mainprice_goal_2016,rudenko_human_2020,sadigh_planning_2018,schwarting_social_2019,kim_anticipatory_2018,dragan_effects_2015,tellex_asking_2014}. From a behavior modeler’s perspective this is great, fueling arguments in papers and funding proposals for the importance of modeling work: To do better HRI [or whatever application one focuses on], we must have better models of behavior! In practice, however, since human cognition and behavior is so endlessly complex, for this applied argument to work, modeling efforts ought to be focused where they are actually useful, which raises the important, yet rarely explicitly addressed question: \emph{What exactly is it that needs modeling, and how accurate do the models need to be?}

In this paper, we will argue that this is a question of general relevance to HRI, and that it is particularly pressing in the context of road vehicle automation. Below, we first provide some background on the use of human behavior models in HRI. We then discuss why models have been useful in these past settings, leading to the suggestion that modelers should explicitly consider the end goal of a \emph{successful human-robot interaction} while scoping their modeling efforts. This leads on to a discussion on how to do this in practice, with concrete examples in vehicle automation, and some thoughts on the role for mechanistic versus machine-learned models of human behavior, as well as suitable next steps for research and development.

\section{Typical uses of human models in HRI}
We adopt an inclusive definition of a \lq model\rq~of human behavior, as any \emph{description} of human behavior, regardless of whether that description is purely qualitative and conceptual or intricately quantitative, and regardless of whether the description suggests underlying mechanisms or is purely phenomenological. 

As for \emph{conceptual models} of human behavior, these have been put to use especially in high-level design of robots, for example by first describing (i.e., modeling) the high-level strategies that humans adopt when handing over objects \cite{strabala_towards_2013} or resolving conflicts in domestic locomotion \cite{babel_step_2022}, and then designing robots which tap into those same strategies. 

More common in HRI, however, is the use of \emph{quantitative models} capable of mathematically describing or predicting some aspect of how humans behave, and/or cognitive mechanisms underlying that behavior.
The most common way of leveraging such models in HRI is to integrate them in the robot's algorithms for perception, planning, or movement. This, in turn, can serve a number of different (sometimes overlapping) purposes:

Some researchers have modeled humans in order to \emph{imitate human behavior}, at some level of abstraction. For example, by formulating a computational model of human-human object handover, one can then use this model directly in a robot controller \cite{huang_adaptive_2015}, and models of human drivers' speed- and lane-keeping can guide the motion control of automated vehicles (AVs) \cite{hajiseyedjavadi_effect_2021}.

Another common approach is to use models that predict observable human behavior from hidden states of the human, in order to permit the robot to \emph{make inferences about human states from observation}. With this type of approach, a robot can for example recognise human movement goals or intentions for short-term body movements \cite{mainprice_goal_2016}
or longer-term locomotion \cite{rudenko_human_2020}. The same general approach can also be applied to infer for example attentional state \cite{sadigh_planning_2018} or social value orientation characteristics of human interaction partners \cite{schwarting_social_2019}. Once the robot has access to estimates about current states of the human, it can adapt its behavior to optimise the interaction accordingly. 

A related type of optimisation instead uses models to \emph{predict future human response to robot behavior}, to steer the interaction toward desirable human states or behaviors. For example, robot behavior in forceful interactions such as collaborative lifting can be controlled to keep human joint load or muscular effort within acceptable ranges \cite{kim_anticipatory_2018}, 
models of how humans interpret robot behavior or utterances can help create more understandable robots, for improved human performance in collaborative tasks \cite{dragan_effects_2015,tellex_asking_2014}, or likely human response to robot actions can be predicted to allow the robot to directly influence human behavior, to achieve an intended interaction outcome \cite{sadigh_planning_2018}. 

Finally, quantitative human behavior models can also be used outside of robot algorithms altogether, to test or benchmark robots in computer simulation. This is less common in general HRI, but is of growing importance in vehicle automation, for large-scale simulated testing \cite{feng_intelligent_2021}, and for assessing AV performance in safety critical situations against the benchmark of an 'attentive human driver performance model' \cite{noauthor_regulation_2021}.

\section{Keeping the end goal in sight}
So why are human behavior models useful in the various examples reviewed above? Simply put, it is because the models help achieve the end goal of a \emph{successful human-robot interaction} (or they allow testing for that end goal). There are many possible metrics by which one might measure interaction success \cite{aly_metrics_2017,hoffman_evaluating_2019}, but we would suggest that all of them can be usefully subsumed under what is sometimes referred to as the three main goals of human factors engineering: \emph{safety, performance, and  satisfaction} \cite{lee2017designing}. 

\begin{figure}
\centering
  \includegraphics[width=\columnwidth]{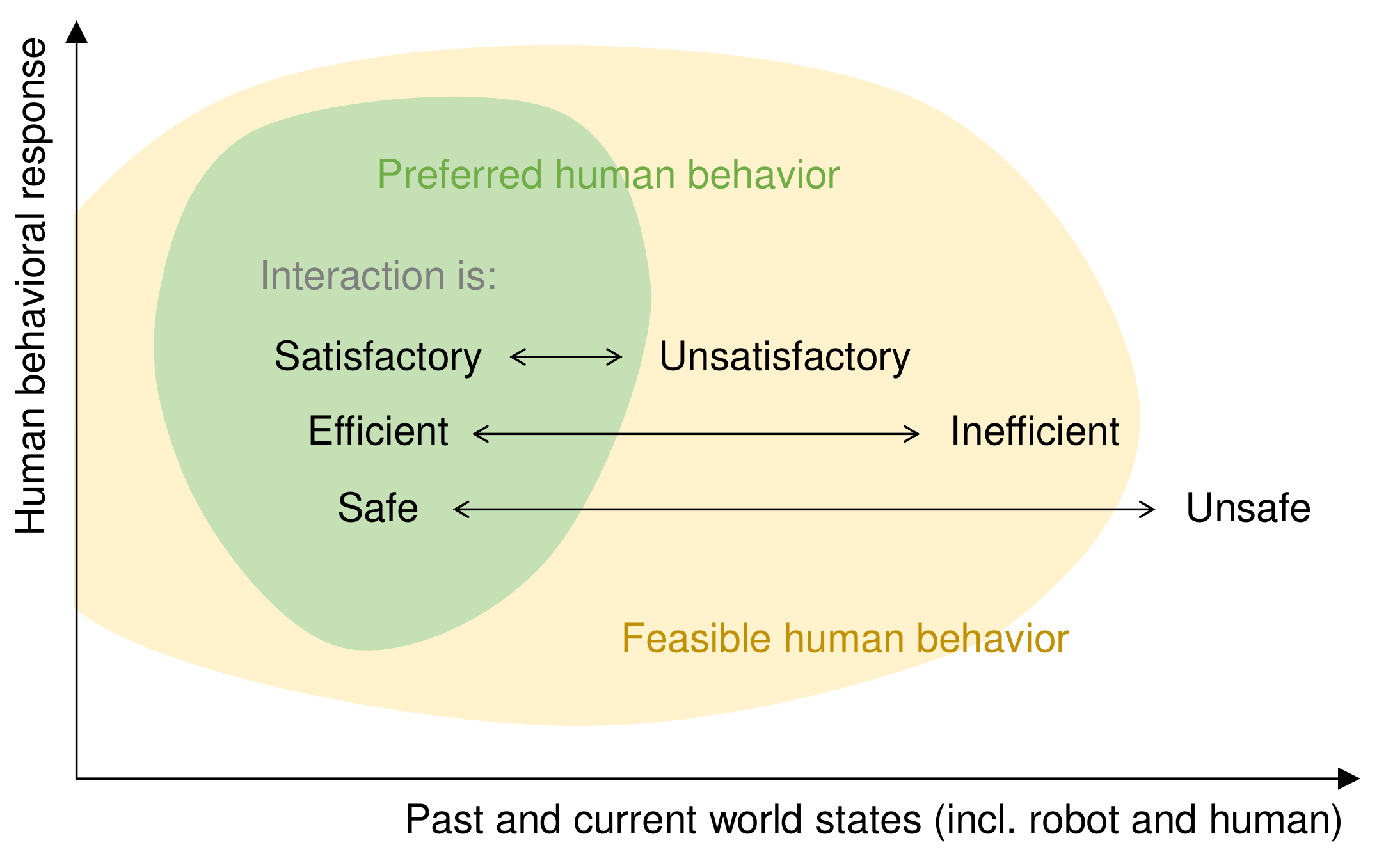}
  \caption{Schematic illustration of preferred and feasible envelopes of task-achieving human behavior in interaction with a robot, and possible impacts on the interaction, if robot behavior is such that the human can't achieve the task within its preferred envelope, or not at all.}
  \label{fig:envelopes}
\end{figure}

It is worth, however, to take a closer look at how, more precisely, the human behavior models help in these respects. We believe that the following roughly captures what is happening here: The reason human behavior can at all be modeled is that it has certain regularities (for example due to stable underlying mechanisms). As illustrated in (the highly schematic) Fig.~\ref{fig:envelopes}, these regularities manifest themselves in the form of constraints on the behaviors expressed by humans during interaction with a robot, both in terms of (i) what behaviors are at all \emph{feasible}, in the sense that they achieve the task at some minimum acceptable level of success and remain available to the human in the given state of the world (e.g., do not require superhuman response times), and (ii) which of these behaviors the human would \emph{prefer} to engage in.  Outside of the preferred envelope, human satisfaction and/or task performance drops, and outside of the feasible envelope, the interaction may fail completely, in some contexts with risks to human safety. 

Under this conceptualisation, the goal of human behavior modeling in HRI becomes one of promoting interaction success, by delineating the feasible or preferred behavioral envelopes of the humans with which the robot is interacting. In principle the model could describe these envelopes explicitly, but the more common approach is instead that the model describes human behavior trajectories, and as long as these model trajectories stay inside of the true behavioral envelopes of the real human, this will still be helpful.



\section{Adapting modeling scope to HRI context}

\subsection{Modeling scope as a context-specific research question}
In any given applied HRI setting, there are still myriad factors and mechanisms which might affect the trajectories and envelopes of human behavior, so still: what do we need to model, and how accurately? Again, we think it is helpful to consider the end goal of interaction success, with the following implication: \emph{The more sensitive interaction success in the given HRI context is to variations in a certain aspect of human behavior or cognition, the more accurate that part of the human model needs to be.} 

In relatively simple interactions this is quite easy to see, especially in hindsight of a successful implementation. For example, in the collaborative tea-making task in \cite{dragan_effects_2015}, quick human understanding of robot motion was key to an efficient interaction, hence an accurate model of human motion understanding was helpful. However, since the human was able to optimise their own object manipulations as they saw fit, a model of human limb movements would not have made any difference to the robot, whereas such a model was key to enabling smooth object handovers in the collaborative dish rack unloading task in \cite{huang_adaptive_2015}.

\begin{figure}
\centering
  \includegraphics[width=0.8\columnwidth]{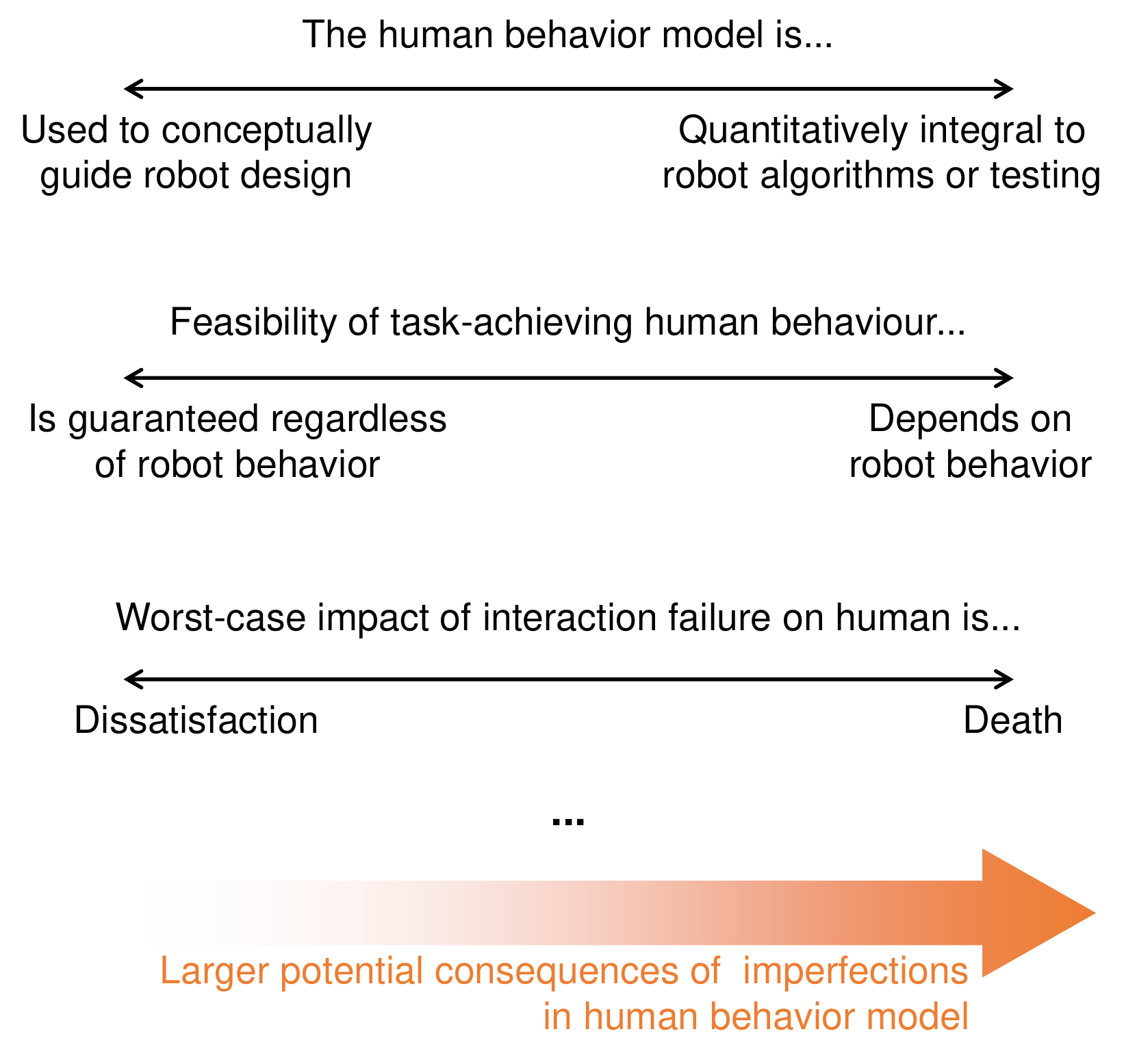}
  \caption{Examples of factors which affect the potential consequence of imperfections in a human behavior model supporting human-robot interaction.}
  \label{fig:consequences}
\end{figure}

Imperfections in these key parts of the human model will be particularly likely to cause the robot to force the human away from their preferred behavioral envelope, degrading the quality of the interaction. The actual consequences of such a degradation depend on the specific HRI context in various ways, some of which are listed in Fig.~\ref{fig:consequences}. For example, if using a conceptual model of human locomotion conflict resolution to guide design of a domestic robot \cite{babel_step_2022}, the result of an incorrect model may be experiment participants which achieve their task in a less satisfying or efficient way, and the need to do another robot design iteration. In contrast, an inaccurate quantitative model of human behavior used in AV algorithms or testing regimes could result in pushing a human road user all the way out of their feasible behavioral envelope, with severe consequences.

Compared to the simple examples given with hindsight above, in a new or more complex HRI context it may initially not be as clear what needs to be modeled. We suggest that in HRI contexts where the potential consequences of model imperfections are large, it is important to consider the question of \emph{what aspects of human cognition and behavior have the biggest impact on interaction success} as a research question in its own right.

\subsection{Answering the research question}
As with any research question, it may be possible to formulate sensible a priori hypotheses about the answer, and perhaps especially so in cases where robots are to be involved in interactions which already take place between humans (such as object handover, collaborative assembly, interaction in road traffic, etc.). In such contexts, we would like to suggest that it may be useful to consider existing empirical research on human-human interactions (as well as introspection, to some degree) to attempt answering the question: \emph{\lq\lq How much could aspect X of human cognition/behavior be simplified in this human interaction, before it yields significantly different interaction outcomes?\rq\rq}

\begin{figure}
\centering
  \includegraphics[width=\columnwidth]{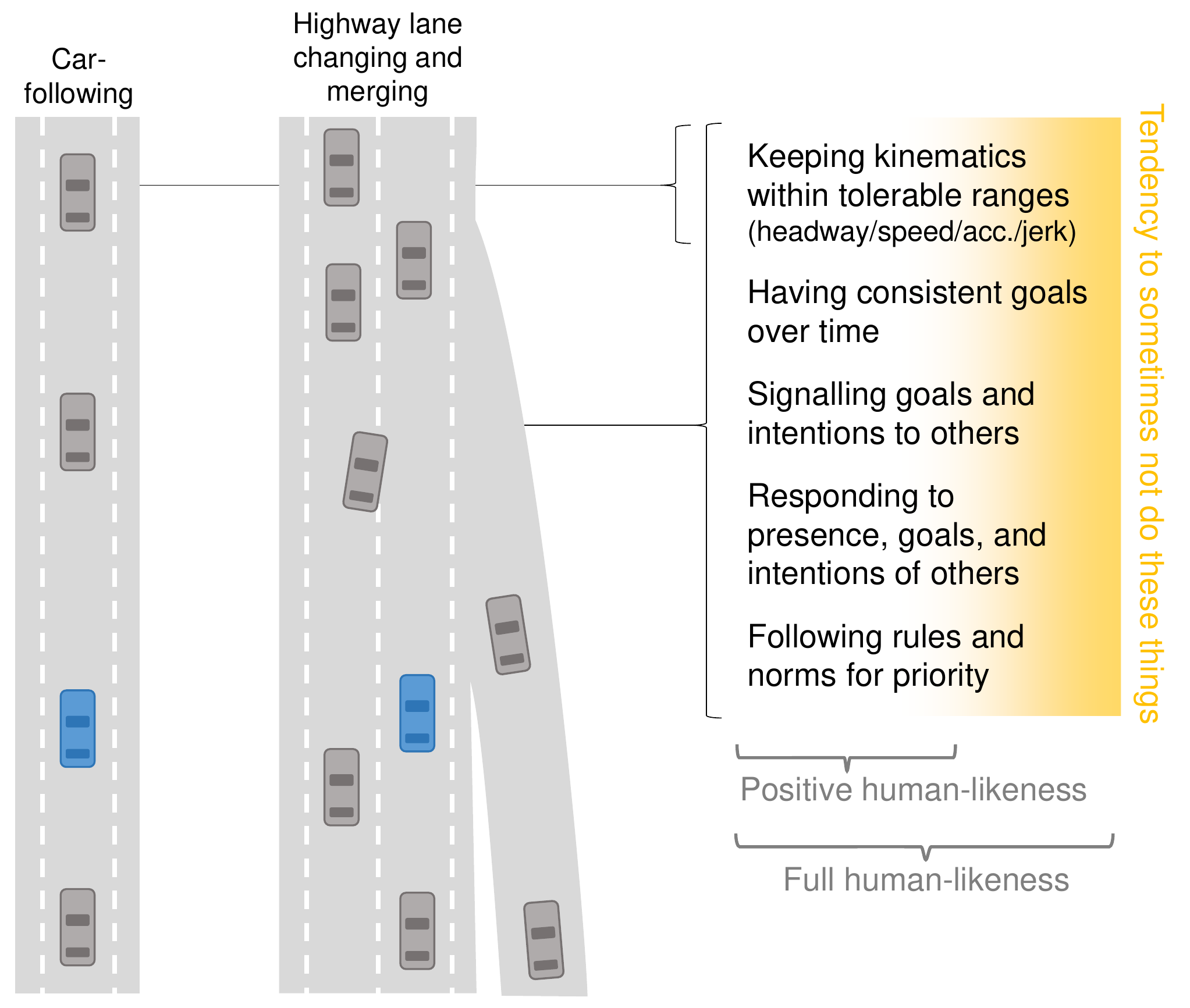}
  \caption{Example road user interaction scenarios, and the human behaviors that a model would arguably need to replicate in order to achieve similar interaction outcomes to human drivers.}
  \label{fig:driving}
\end{figure}

To clarify what we mean here, let us consider the two examples of road user interaction shown in Fig.~\ref{fig:driving}. Imagine that you are driving the blue car, and that the surrounding gray vehicles are driven either (i) all by actual human drivers, or (ii) all by models of human drivers. For the simpler case of pure car-following (left in the figure), there exist many models of what perceptual quantities and motor strategies human drivers use to achieve this type of longitudinal control \cite{saifuzzaman_incorporating_2014}, but these details may not be of great importance for your interaction with the surrounding vehicles. As long as the gray vehicles keep their kinematics within typical ranges for human drivers, keeping speed up to make progress, avoiding excessive deceleration and so on, you would possibly not perceive the interaction with these model-driven vehicles as significantly different from typical interactions with human-driven vehicles. These simple models could thus be regarded as human-like in a \emph{positive} sense, and could be perfectly useful for some HRI contexts, e.g., to generate human-like automated car-following.

However, as also indicated in Fig.~\ref{fig:driving}, in the same simple car-following scenario there may be a need to go beyond this positive human-likeness, and also consider the human tendency to sometimes \emph{not} keep kinematics within typical human ranges. If, as you drive the blue vehicle, your gray lead vehicle \emph{never} exhibits late reactions or excessive decelerations, then your interaction with it will be safer than with a human lead, so in this sense the simplified model does yield a significantly different interaction outcome. This can be important if the intended use of the model in an AV is to predict likely future behavior of human-driven lead vehicles, where too benign model behavior could make for an AV which is not sufficiently defensive. Again, however, it seems unlikely that very detailed models (e.g., of human visual time sharing or exact decision-making mechanisms) would be needed to achieve this fuller human-likeness. Quite possibly, it could be enough to make sure that the ranges of kinematics covered by the model, for example in the form of probability distributions, also include these more uncommon extremes \cite{feng_intelligent_2021}.
 
If we now instead consider the other scenario shown in Fig.~\ref{fig:driving}, including also lane changes and merges, the list of behavioral phenomena to consider grows rapidly. Here, it seems unlikely that your interactions with models would have similar outcomes to those with humans if the model drivers did not have consistent goals over time, did not signal near-term intentions explicitly (i.e., turn indicators) and/or implicitly (e.g., lane positioning), or did not respond to such indications from you to also accommodate your goals, to the extent expected given local rules and norms. While progress is being made in this area \cite{schwarting_social_2019,kang_repeated_2020}, it remains an open question how to model these types of human road user behaviors, which may involve cognitive mechanisms such as theory of mind and social value orientation in decision-making, and crucially, the listing of aspects of behavior in Fig.~\ref{fig:driving} is still just an a priori hypothesis: \emph{To know which of these behaviors or mechanisms actually have the most impact on interaction success, dedicated empirical and modeling work will be needed}. And this is just to achieve human-likeness in the positive sense, without yet considering how to account for human shortcomings in these situations. 

The examples above illustrate what we mean by formulating hypotheses about which aspects of human behaviour have an impact on interaction success in a given task context, based on reasoning from current human interactions. But especially the latter example also illustrates how this research question may, especially for more complex HRI contexts, require iterative modeling and empirical work to reach an answer.

\section{A note on machine-learned models}

Note that we haven't argued above that human behavior models need to be mechanistic. Data-driven, machine-learned modeling of human behavior is achieving increasingly impressive results, and will likely be key to many application areas, not least vehicle automation \cite{suo_trafficsim_2021}. According to what has been said here, as long as these models accurately capture those aspects of human behavior which most affect interaction outcomes, all is good. However, it is important to note that the machine-learned models will not by themselves tell us \emph{which those key aspects of behavior are}, so the research question emphasised in this paper remains unanswered. In other words, until we have done the research to identify the aspects of human behavior to which a given interaction is most sensitive, we have no way of knowing whether the machine-learned models are fit for purpose. We think this insight highlights an important weakness in purely machine-learned modeling for HRI, but it also clearly indicates a path to addressing it. 

\section{Next steps}

In our view, the most important next step in this area is to increase the focus on the key research question identified here: To what aspects of human behaviour is human-robot interaction success most sensitive? This will require research leveraging both naturalistic and controlled empirical studies, to understand what humans do, and how, in order to achieve human-human and human-robot interactions that are successful (by some metrics, which may also need further development). These aspects of human behaviour can then be targeted in modeling. The resulting improved human models can also be put to use, to further address the same research question, for example using ablation methods: Both in pure model simulation studies and in controlled studies where humans interact with model-controlled robots or with virtual humans, what model assumptions or capabilities make the biggest difference to interaction success?

\section{Conclusion}
We have argued that for applied HRI purposes, human behavior modelers should focus on those aspects of behavior to which interaction outcome is the most sensitive, and the question of which those aspects of behavior are should be considered an important research question in its own right. Answering this question is not trivial, and will require targeted research of its own, but the answers to it will tell us what needs modeling, and what to look for when testing (not least machine-learned) models. The exact requirements on model accuracy will vary between HRI contexts, because the potential knock-on consequences of human model imperfection varies, with vehicle automation as a clear example of an application area warranting particularly high standards of accuracy for human behavior models.


\section*{Acknowledgment}

This work was supported by the UK Engineering and Physical Sciences Research Council under grant EP/S005056/1.



\bibliographystyle{IEEEtran}
\bibliography{references,nourl}
%
%
%


\end{document}